\documentclass[letterpaper, 10 pt, conference]{IEEEconf}  % Comment this line out if you need a4paper

\IEEEoverridecommandlockouts                              % This command is only needed if 
\usepackage{graphicx}
\usepackage{amssymb}
\usepackage{multirow} 
\usepackage{lipsum}
\begin{document}

\title{\LARGE \bf
Self-supervised 6-DoF Robot Grasping by Demonstration via Augmented Reality Teleoperation System
}

\author{
Xiwen Dengxiong \\
\small{Rochester Institute of Technology} \\
\small{Rochester, NY, USA} \\
\small{sd6384@rit.edu}
\and
Xueting Wang \\
\small{Rochester Institute of Technology} \\
\small{Rochester, NY, USA} \\
\small{xw4860@rit.edu}
\and
Shi Bai \\
\small{Figure AI} \\
\small{Sunnyvale, CA, USA} \\
\small{baishi.bona@gmail.com}
\and
Yunbo Zhang \\
\small{Rochester Institute of Technology} \\
\small{Rochester, NY, USA} \\
\small{ywzeie@rit.edu}
}

% \author{\IEEEauthorblockN{Xiwen Dengxiong}\\
% \IEEEauthorblockA{
% % \textit{Dept. of Computing and Information Sciences} \\
% \small{\textit{Rochester Institute of Technology}}\\
% Rochester, NY, USA\\
% sd6384@rit.edu} \\
% \and
% \IEEEauthorblockN{Xueting Wang}\\
% \IEEEauthorblockA{
% % \textit{Dept. of Industrial and Systems Engineering} \\
% \small{\textit{Rochester Institute of Technology}}\\
% Rochester, NY, USA\\
% xw4860@rit.edu} \\
% \and
% \IEEEauthorblockN{Shi Bai}\\
% \IEEEauthorblockA{
% \small{\textit{Figure AI}} \\
% Sunnyvale, CA, USA \\
% baishi.bona@gmail.com}
% \and
% \IEEEauthorblockN{Yunbo Zhang}\\
% \IEEEauthorblockA{
% % \textit{Dept. of Industrial and Systems Engineering} \\
% % \textit{School of Information (affiliated)} \\
% \small{\textit{Rochester Institute of Technology}}\\
% Rochester, NY, USA\\
% ywzeie@rit.edu} \\
% }

\maketitle

%%%%%%%%%%%%%%%%%%%%%%%%%%%%%%%%%%%%%%%%%%%%%%%%%%%%%%%%%%%%%%%%%%%%%%%%%%%%%%%%
\begin{abstract}

Most existing 6-DoF robot grasping solutions depend on strong supervision on grasp pose to ensure satisfactory performance, which could be laborious and impractical when the robot works in some restricted area. 
To this end, we propose a self-supervised 6-DoF grasp pose detection framework via an Augmented Reality (AR) teleoperation system that can efficiently learn human demonstrations and provide 6-DoF grasp poses without grasp pose annotations. Specifically, the system collects the human demonstration from the AR environment and contrastively learns the grasping strategy from the demonstration. For the real-world experiment, the proposed system leads to satisfactory grasping abilities and learning to grasp unknown objects within three demonstrations.

\end{abstract}

\begin{figure*}
    \centering
    \includegraphics[width=\textwidth]{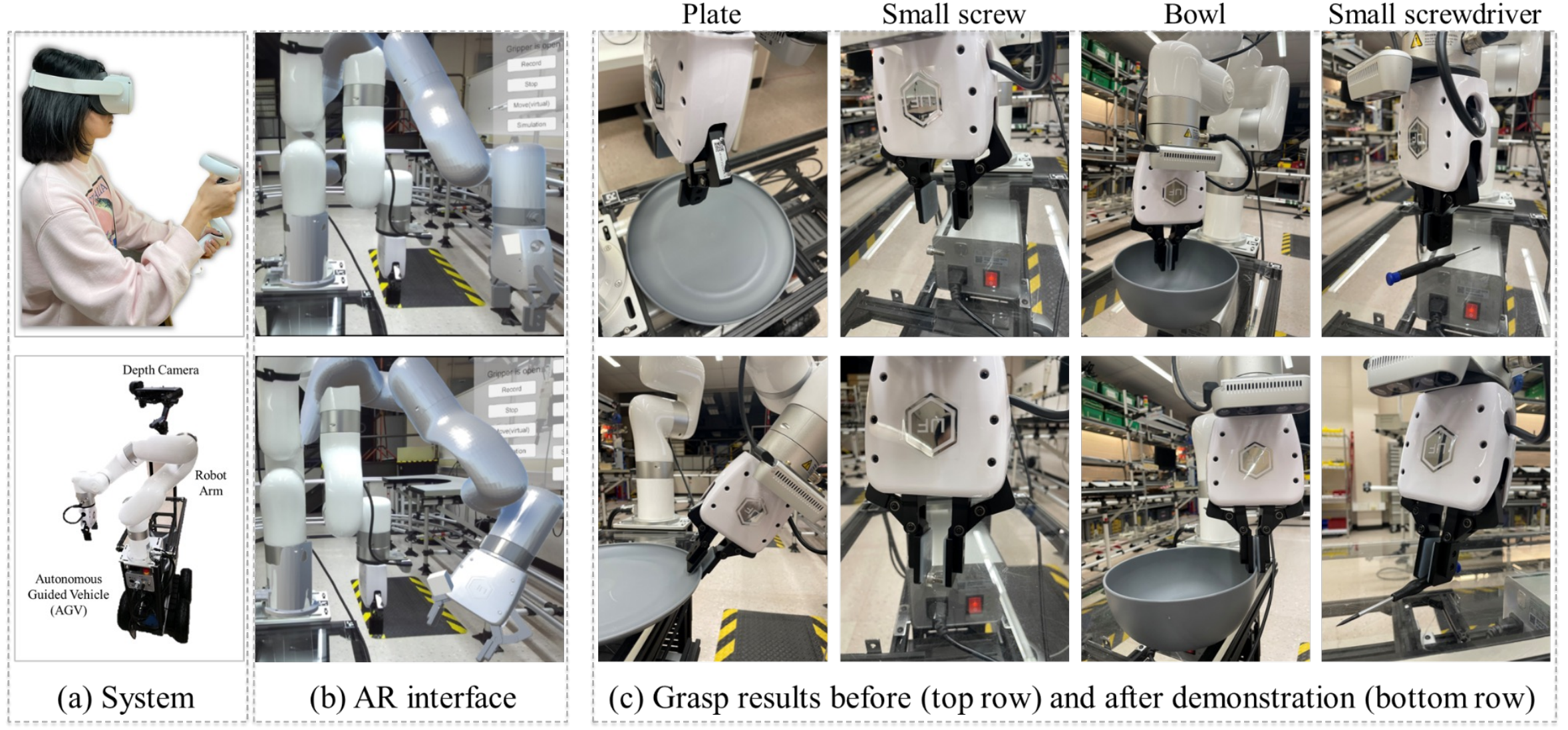}
    \caption{6-DoF grasp pose for unknown objects. Red gripper shows the grasp pose. The first row illustrates the generated grasp poses without human demonstration. The second row shows the final grasp pose after learning demonstrations.
    }
    \label{task}
\end{figure*}

\section{Introduction}

Unknown object grasp pose detection aims to generate 6-DoF grasp poses for unknown objects. It is a significant task in real-world applications because robots may need to grasp unknown objects under extreme environments, such as nuclear plants \cite{moore1985robots}, underwater environments \cite{wang2022sliding}, and outer space \cite{wang2020event}. 

However, detecting grasp pose for unknown objects is challenging because 6-DoF grasp pose needs to be learned from accurate grasp pose annotations. Although the problem can be solved by manually providing accurate grasp poses, physically annotating the robot grasp pose in an extreme environment is impractical. Moreover, limited grasp pose annotations may impact the performance of the grasp pose detection method \cite{wang2021graspness,Dengxiong_2023_WACV}. Therefore, a solution that can efficiently learn to grasp unknown objects in remote environments is required.

In this paper, we propose a self-supervised 6-DoF robot grasp pose detection framework via an Augmented Reality (AR) teleoperation system that can efficiently learn to grasp unknown objects through human demonstrations in the AR environment. Specifically, the framework first generates the initial grasp pose to detect the location of the object based on the image segmentation method. Although the initial grasp pose cannot perfectly grasp the object, it still can roughly illustrate the location of the detected object for further adjustment.
Then, the AR teleoperation system collects a few human demonstrations through AR software, which enables humans to provide accurate demonstrations remotely. Human demonstration contains waypoints from the start pose to the grasp pose, and the RGB-D image at the start pose. After that, the framework utilizes a point clouds-based contrastive learning method to learn the hidden morphology representation and evaluate the similarity between the detection object and the object demonstrated by humans. Finally, the framework generates 6-DoF adjustment to improve the initial grasp pose to the 6-DoF grasp pose for unknown objects.

As shown in Fig \ref{task} section c, the framework first detects the initial grasp pose for four objects, which cannot grasp the unknown objects, e.g., plates, small screws, bowls, and small screwdrivers. In Fig \ref{task} section b, The AR interface is used to provide waypoints remotely. After learning several demonstrations, the framework can generate an adjusted 6-DoF grasp pose for unknown objects (Fig \ref{task} section c).
Compared with the existing supervised robot grasping works \cite{fang2023anygrasp,wang2021graspness}, our system can generate the 6-DoF grasp pose and grasp the object without grasp pose supervision because the contrastive learning model learns the hidden morphological representation and can be transferred to adjust the 6-DoF grasp pose based on demonstration. Different from robot imitation work \cite{sermanet2016unsupervised,asfour2008imitation}, our framework can efficiently use human demonstrations as prompts to generate the 6-DoF grasp pose and learn the waypoints from demonstrations. Additionally, the system is designed as a distributed system to allow remote control by the user and analyze the RGB-D information in different platforms.
Our contributions are summarized as follows:
\begin{itemize}
    \item We propose a novel self-supervised demonstration learning framework to contrastively learn the grasp pose from human demonstration and generate 6-DoF grasp poses.
    \item We propose a 6-DoF grasp pose detection method that efficiently learns the 6-DoF adjustment solution within several demonstrations (3 times in our experiment).

\end{itemize}
\section{Related Work}

\paragraph{Vision Based Robot Grasping}
The grasp pose detection problem can be formulated as detecting the grasp pose and moving the end effector to the detected grasp pose. 
Earlier methods consider learning the correlations between multiple representations from detection and the final grasp pose \cite{wen2022catgrasp}. The input representations include point cloud representations \cite{ni2020pointnet++, li2021simultaneous}, rectangle representations\cite{jiang2011efficient}, and grasp quality maps representations\cite{choi2018learning, wang2020q, wu2020grasp}.
% These methods mainly reduce the 6-DoF grasp pose into generating 4-DoF grasp poses on the camera plane. 
However, these methods may ignore the key poses on the target object and fail, e.g., flat plates. 
Some approaches consider RGB-D inputs\cite{gou2021rgb,ni2020pointnet++} to regress the 6-DoF grasp pose, and other works \cite{fang2020graspnet, wang2021graspness} propose to use the cloud points to generate redundant grasp poses for the objects. Specifically, Fang \textit{et al.} \cite{fang2020graspnet,fang2023anygrasp} Generate abundant grasp poses based on the large-scale dataset with analytic labels on grasp pose, and Wang \textit{et al.}\cite{wang2021graspness} eliminate the unfeasible grasp poses by proposing graspness score to evaluate the generated grasp pose. Although the grasp pose generating module shows high accuracy in grasping seen and similar objects based on the large-scale dataset, the performance is limited when only a few grasp pose annotations are given. In contrast, our framework proposes to acquire the grasp strategy without grasping annotations and use the morphology representations to learn the human demonstrations selectively. This allows our framework to enhance the grasping ability for unseen objects after learning a few human demonstrations.

\paragraph{Human Demonstration Learning via Teleoperation System}

The robot teleoperation system aims to enable human operators to manipulate the robot in different working spaces \cite {guo2019scaled,liang2022learning,wen2022you,liang2021vision}, which provides an efficient solution to transfer human instructions to the robot system.
Compared with teleoperation solutions in recent works \cite{baklouti2021improvement,young2020review,gliesche2020kinesthetic,clark2019role,aggravi2021haptic,mizera2019evaluation,handa2020dexpilot,sivakumar2022robotic}, AR-based teleoperation system \cite{pan2021augmented, qin2023anyteleop,arunachalam2023holo,wanginvestigating,xiang2020sapien,todorov2012mujoco,gan2020threedworld} is notably beneficial because it provides fully immersive or context-rich interactions with the remote environments and enables easy and accurate human demonstrations collecting feature. Specifically, Pan \textit{et al.} \cite{pan2021augmented} proposes to use the RGB-D image and pose teaching device to enhance the interaction between human and robot interaction. Li \textit{et al.} \cite{li2022ar} collect human demonstrations and learn robot manipulation. Although these frameworks have better interaction experience, the robot system can barely learn useful grasp strategies through limited human demonstration. 
Some imitation learning methods \cite{sermanet2016unsupervised, calinon2007incremental} propose to learn the human demonstration through reinforcement learning. These methods need to learn the reward policy of grasping when demonstrations are given.
In our works, we propose to learn the point clouds of the objects in a contrastive way. Demonstrations are used as prompts to enable the 6-DoF grasp pose adjustment. This allows the system to efficiently generate 6-DoF grasp poses by comparing the morphology representations.

\begin{figure*}
    \centerline{\includegraphics[width=\linewidth]{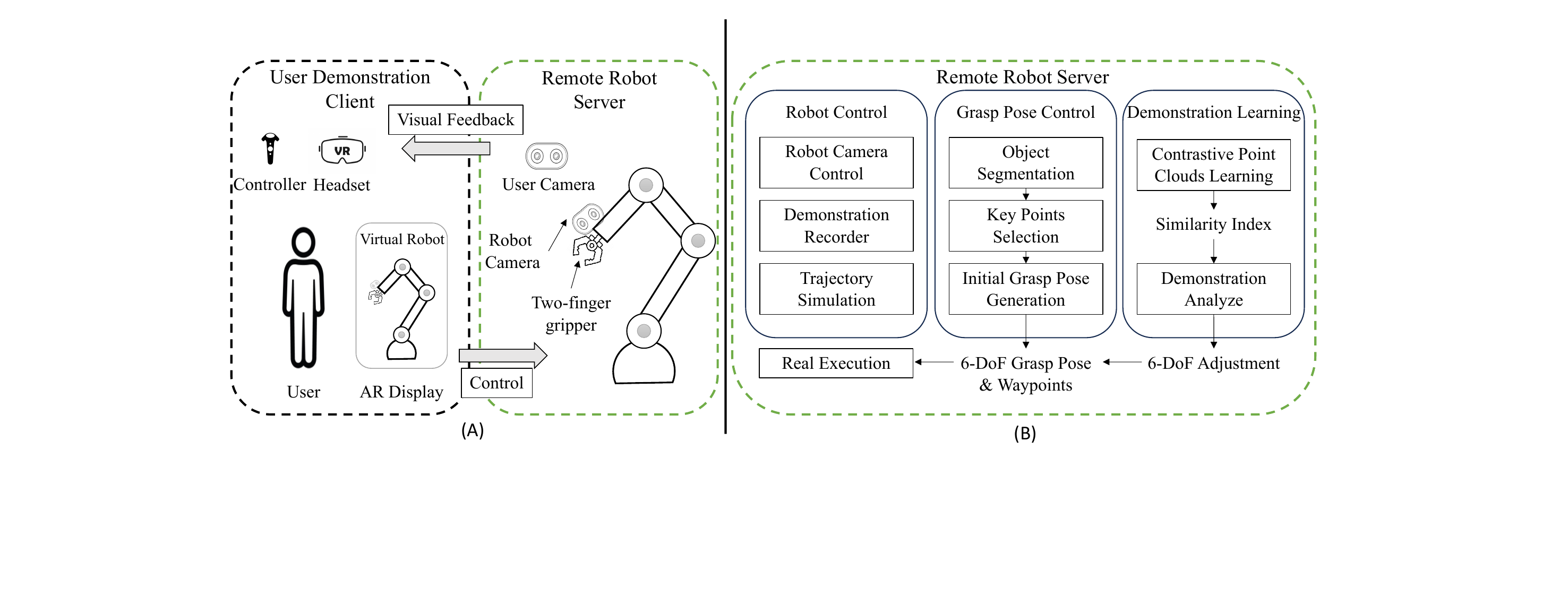}}
    % \captionsetup{aboveskip=3pt}
    \caption{ \small{System Overview. Section A illustrates the system design. The user camera is used to render the remote environment in the AR display. Users may use the AR software to control the remote robot. The robot camera can collect RGB-D images for robot grasping. Section B shows the architecture of the remote robot server, including Robot Control, grasp pose control, and demonstration learning. }
    % The robot control module directly controls all hardware on the remote side. 
    % The grasp pose control module generates 4-DoF initial grasp pose based on the RGB-D image, and the demonstration learning module adjusts the grasp pose and generates 6-DoF final grasp pose. 
    }
    \label{framework}
\end{figure*}
\section{System Overview}

The proposed AR teleoperating system contains an immersive AR environment for users to provide demonstrations \cite{qiu2020human}, a grasp pose control module, and a demonstration learning module. It is designed as a distributed system that includes a user demonstration client and a remote robot control server.
Fig \ref{framework} illustrates the proposed system and the key components in the system. We realize the following features based on the proposed system.

\begin{itemize}
    
    \item \textit{Initial grasp pose control}
    The AR teleoperating system can find the target objects and generate a 4-DoF initial grasp pose based on the segmentation masks and the depth map from the RGB-D camera.
    \item \textit{Collecting user demonstration in augmented reality}
    Users can provide grasping instructions by demonstrating the grasping procedure in the AR environment.
    \item \textit{Demonstration learning}
    The system records user demonstrations and the corresponding RGB-D camera input as prompts to generate 6-DoF grasp pose and waypoints based on user demonstrations.
\end{itemize}

% We compared our proposed system with other robot teleoperating system in two dimensions (1) user operating requirements (2) user instruction case

\subsection{System Design}
In the AR teleoperation system, the remote robot server directly links with two cameras and a robotic arm. The user client connects the VR headset and controllers. As shown in Fig \ref{framework}, the robot server is designed for collecting visual data, finding the grasp pose, and providing waypoints for grasping. Both cameras are RGB-D cameras, and the robot camera is located on the end-effector of the robotic arm.
The visual data from the user camera can be rendered in the AR environment, while visual data from the robot camera are used for generating grasp pose. Users can use the AR software to control the virtual robot to visualize the demonstration and manipulate the real robot to grasp objects. The user demonstration contains object point clouds and the waypoints that move the end effector from the start pose to the grasp pose. If the computing resources of the robot computer are limited and cannot process the complicated segmentation and 6-DoF grasp pose adjustment workload, the grasp pose control module and demonstration learning module can be executed in another powerful remote server. Overall, users can directly use the controller to provide demonstrations in the user client and control the remote robot server.

\section{User Demonstration Client}

To visualize the user demonstration feature in the immersive AR environment \cite{cao2023real}, we implement demonstration software based on Unity 3D. 
Inspired by \cite{zhang2023educational}, we integrate the Oculus VR headset and controller in Unity to render the visual data from the user camera and control components in the AR environment. The user camera is physically connected to the robot computer and transfers the visual data to the user computer through the internet to realize remote teleoperation tasks.

In the demonstration software, we import a virtual robot with a virtual two-finger gripper. The virtual robot is calibrated with the real robot in the world coordinate.
There is another virtual gripper in the scene indicating the pose of the end-effector recorded by the software. During the demonstration process, the user needs to use the controller to drag the virtual gripper. The virtual gripper can follow the user controller and move in the AR environment. Here, we provide two gripper following solutions. Users can click the button on the controller to decide pure position following or position and orientation following. Besides the user can also control the gripper status by pressing the button on the controller. There are several virtual buttons in the panel of the AR environment that offer significant features for controlling the virtual robot and the real robot. Specifically, the real robot can be driven by initialize, search, and execute buttons. The back button can cancel the recorded virtual trajectory. Move and simulate all buttons only manipulate the virtual robot based on the recorded trajectory. We connect the software and the robot control system through the ROS-TCP connector \cite{TCPconnector} library and enable the information transfer feature.

\section{Remote Robot Server}

The remote robot server is proposed to generate grasp poses and control the robot based on the RGB-D data from the robot camera. 
It contains three modules (Fig \ref{framework} (B)): (1) Robot control module, which enables teleoperation feature on the real robot and the visualization of the robot movement in the AR environment. (2) Grasp pose control module, which generates the grasp pose without training on grasp pose annotation. (3) Demonstration learning module, which learns the correlation between demonstrations and generates the adjustment on trajectory and final grasp pose. We integrate the robot control module with ROS and realize camera control, record demonstrations, and trajectory simulation features.

\subsection{Grasp Pose Control Module}

The grasp pose control module aims to generate the initial grasp pose without human demonstrations or grasp pose annotations. Given the RGB-D image from the robot camera, the module first segment the RGB image and select the key points from the detected masks. The module then generates 4-DoF grasp poses in the world coordinate with transformation and rotation on the yaw axis.

\paragraph{Object Segmentation} The object segmentation process aims to segment objects from the RGB frame. In the initial grasp pose generating module, we utilize the FastSAM \cite{zhao2023fast}, a lightweight segmentation model that only needs an RGB frame as input. The FastSAM model can generate potential object masks and detection bounding boxes for objects.

\paragraph{Key Points Selection}
The key points selection aims to select the most suitable grasp point based on the segmentation masks and the bounding boxes from the RGB frame. We select five key points for each object and use the pixel position to calculate the confidence of the key points. Five key points contain one point at the center of mass of the segmentation mask, one point located on the mask and has the closest distance from the bounding box center, and the other three points are located on the outline of the segmentation mask. 
% The confidence of the center of mass point is equal to half of the mask confidence, while the closest distance from the bounding box center uses half of the bounding box confidence. The rest of the points have equal confidence, and the five-point confidence summation is equal to one.

\paragraph{Initial Grasp Pose Generation}
The initial grasp pose output consists of translation at three axes and the yaw rotation. To calculate the yaw rotation, we use the center of mass point as the center point and calculate the bounding rectangle. The shorter edge of the bounding rectangle is selected to calculate the yaw rotation. After that, the pixel positions of five key points are utilized to calculate the camera position. The pixel position needs to be transformed to the grasp position. We first align the RGB frame and the depth map and get the camera intrinsic to transform pixel position to camera position. Then, we calibrate the camera and the robotic arm gripper and get the eye-in-hand calibration matrix.
Finally, we transform the pixel position to the 4-DoF grasp pose.
% for the robotic arm based on the position of the robot arm end effector.
% The initial grasp pose position is the highest confidence key point position.
% The generated initial grasp pose is the grasp pose of the grasping task. 
% The initial grasp trajectory includes four key poses. The detection pose is the pose when camera captures the RGB-D frame. The pre-grasp pose is the pose that adjusts yaw rotation. The final grasp pose refers to the pose when the gripper closes. Place pose is the pose when the gripper opens and places the object.

% \begin{figure}[h]
%     \centerline{\includegraphics{contrastive.pdf}}
%     % \captionsetup{aboveskip=3pt}
%     \caption{\small{}}
%     \label{contrastive}
% \end{figure}
% \section{System Overview}
\subsection{Demonstration Learning Module}
\begin{figure}
    \centering
    \includegraphics[width=\linewidth]{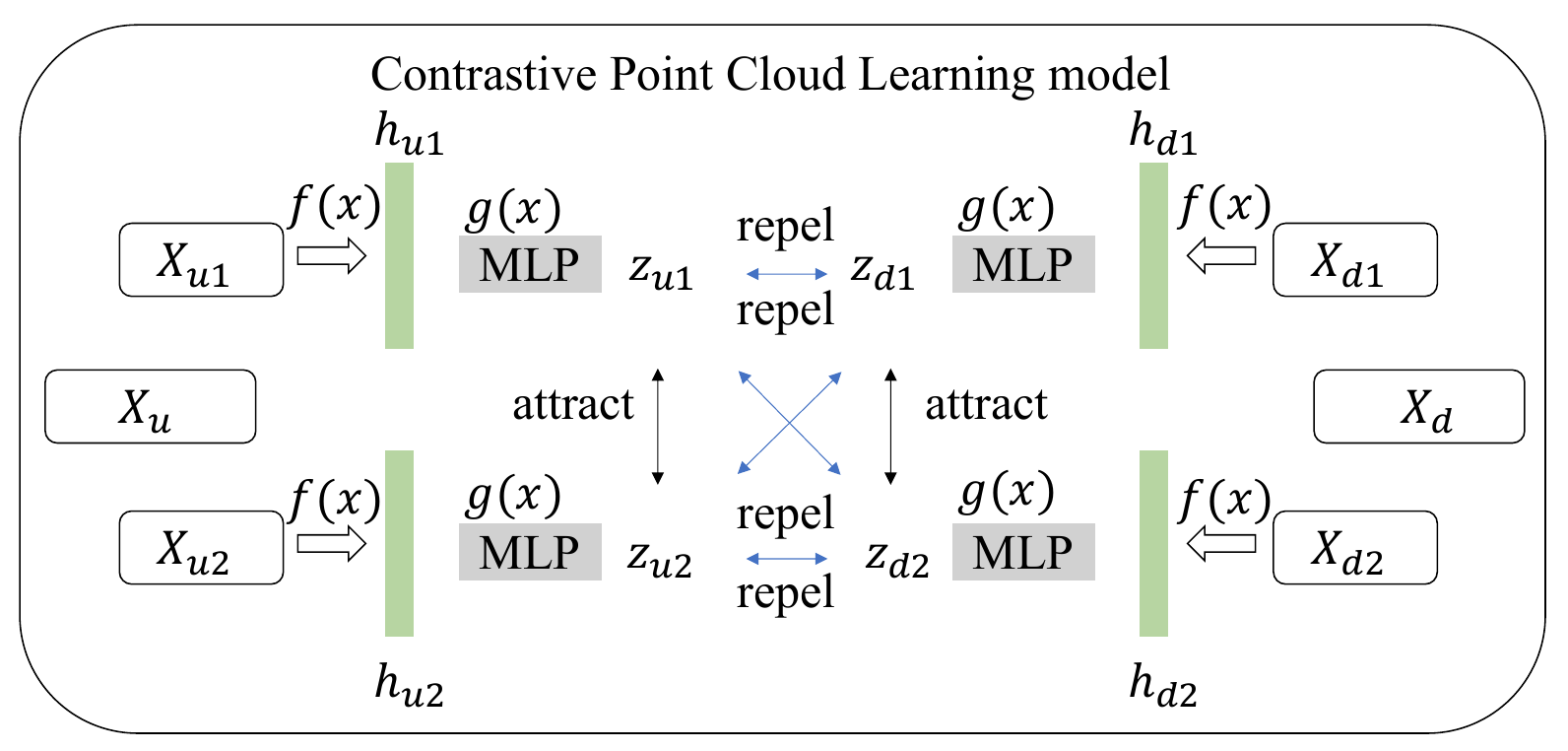}
    \caption{ \small {The contrastive point cloud learning model first augments the input and extracts features through projector $g(f(\cdot))$. }
    }
    \label{contrastive}
\end{figure}

The purpose of the demonstration learning module is to define which object should learn the demonstrations from humans. 
Since human demonstrations are input in the demonstration learning module as prompts, 
the system needs to recognize objects that have similar morphology representation with the demonstrated objects and learn the corresponding human demonstration. 
According to the previous 6-DoF grasp pose generating works \cite{newbury2023deep,breyer2021volumetric,wu2020grasp,zhao2021regnet,wen2022catgrasp}, the 6-DoF grasp pose is mainly influenced by the point clouds of target object. In the demonstration learning module, we use a self-supervised method to contrastively learn the hidden representation of the object point cloud. The Demonstration learning module contains (1) a contrastive point cloud learning model, and (2) a demonstration analyzing section. The output of the module is the 6-DoF grasp pose adjustment.

\paragraph{Contrastive point cloud learning}
The contrastive point cloud learning model aims to separate the hidden point cloud representation in a self-supervised manner. 
% Similar hidden point cloud representations indicate similar object shapes. 
The output of the model is the similarity index.
Inspired by the SimCLR \cite{chen2020simple} that separates the hidden representation of RGB image, we propose to contrastively learn the hidden representation of point cloud information by maximizing the agreement between augmented point clouds in the latent space. 
Since we do not have the grasp labels for the training point clouds, each point cloud sample needs to be augmented into two point clouds to enable comparison between the same objects and different objects in the latent space. 

Specifically in Fig \ref{contrastive}, during the training process, the point cloud sample $x_u$ is defined as the original sample. Firstly, the original sample augments into two samples through jittering, resizing, or flipping, e.g., $x_u$ to $x_{ui}, x_{uj}$. Three augmented methods are randomly selected to increase the robustness of the model.
Secondly, the augmented samples are extracted by encoder $f(\cdot)$ to get the representation vectors $h_{ui},h_{uj}$ of the augmented samples. In this paper, we adopt the ResNet \cite{he2016deep} as encoder and $h$ is the output after the average pooling layer. After that, the representation vector $h_{ui},h_{uj}$ are projected by MLP $g(\cdot)$ to map the representation to the latent space. The projected representation $z_{ui}$ and $z_{uj}$ is a positive pair, and the task of contrastive loss is to identify $x_{uj}$ in a set of positive pair$\{x_{uk}\}_{k\not =i}$ for $x_{ui}$.
Assuming we have $N$ original sample in a batch, we get $2N$ augmented samples. Given a positive pair in the batch, the other $2N-1$ samples are negative samples. The contrastive NT-Xtent loss for $x_{ui}, x_{uj}$ can be formulated as follows.

% \begin{equation}
%     \mathcal{L}_{i, j} = -\log \frac{\exp\left(\frac{\operatorname{sim}(z_{ui}, z_{uj})}{\tau}\right)}{\sum_{k=1}^{2N} \mathbf{1}_{[k \neq i]} \exp\left(\frac{\operatorname{sim}(z_{ui}, z_{uk})}{\tau}\right)}
% \end{equation}

\begin{equation}
\mathcal{L}_{i, j} = -\log \frac{\exp\left(\frac{sim(z_{ui}, z_{uj})}{\tau}\right)}{\sum_{k=1}^{2N} \mathbf{1}_{[k \neq i]} \exp\left(\frac{sim(z_{ui}, z_{uk})}{\tau}\right)}
\end{equation}

where $sim(z_{ui},z_{uj})$ is the cosine similarity and $\tau$ is the temperature parameter. $\mathbf{1}_{[k \neq i]}$ is an indicator function equal to 1 only if $k \neq i $.
In this paper, we use the ShapeNet \cite{chang2015shapenet} point cloud dataset to pretrain the model and update the non-linear projector $g(\cdot)$. 
% During the training process, we do not use the label.

\paragraph{Demonstration Analyzing Section}
% This section illustrates how to generate 6-DoF adjustment based on the similarity index and corresponding demonstration.
Before the demonstration starts, the system will record the RGB frame and the depth map of the experiment surroundings. For the detected objects, the system first transforms the detected masks into point clouds based on the camera coordinate and normalizes the point clouds. Then, the system can generate the initial grasp pose for detected objects. 
When the demonstration starts, the AR software can record a set of waypoints from the start pose to the grasp pose. To avoid the display error in the AR software and get the accurate demonstrated grasp pose, we fine-tune the position of the grasp pose. Each demonstrated sample contains the point clouds and waypoints.

During the experiment, some objects may be demonstrated multiple times. If the demonstrated samples are recording point clouds for the same object, they belong to the same category.
Assuming the software records waypoints from the human demonstration (start pose to grasp pose), the system needs to output the same number of key points from the start pose to the grasp pose for target objects.
There are two scenarios. If the system can generate the initial grasp pose for the demonstrated object, the system can adjust both position and orientation based on the demonstration. If the system cannot generate the initial grasp pose, we only learn the orientation adjustment.
To ensure the number of key points in the initial grasp trajectory is equal to the number of waypoints in the human demonstration, we separate the initial trajectory evenly into $M$ points.
For each segmentation, the system can prepare point clouds for segmentation masks. The prepared point clouds first compare all recorded demonstration point clouds and get the similarity between all demonstrated categories. If there are multiple demonstrated samples for one category, we use the demonstration information with maximum similarity as the similarity index $i_s$ for the category. For each demonstration, we first analyze the recorded waypoints. After finding the center point of the point clouds, we generate four pseudo grasp poses 
% $\hat{P}=(\hat{p_1},\hat{p_2},\hat{p_3},\hat{p_4})$ 
around the outline of the object based on the demonstrated grasp pose. The pseudo grasp poses have the same distance from the center point of the object. The distance from the center point to the pseudo grasp pose
% $\hat{p_i}$ 
can be resized based on the scaling factor between the demonstrated object and the detected object.
There are two thresholds indicating learning rotation $t_r$ and learning translation $t_l$. If $i_s > t_r$, the system utilizes the generated pseudo grasp pose as the final grasp pose.
% To complete the trajectory planning procedure, we choose the first pseudo grasp pose that is located in the segmentation result. 
If $i_s>t_l$, the system will use the demonstrated grasp pose provided by human.

\begin{figure}
    \centering
    \includegraphics[width=\linewidth]{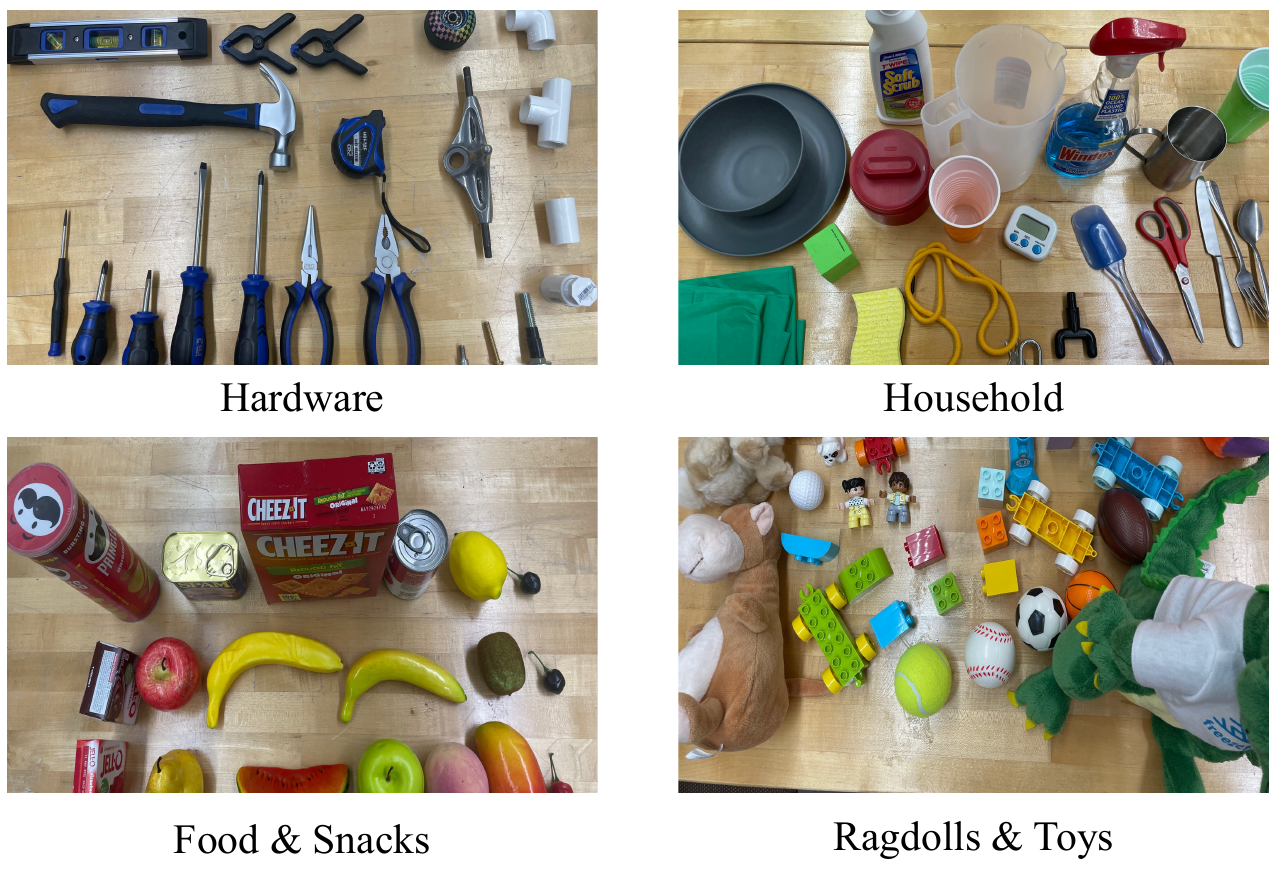}
    \caption{\small{Example objects in hardware, household, food, and toys.}
    }
    \label{objects}
\end{figure}
\section{System Evaluation}
% To verify the performance of our grasp solution and demonstration learning, we integrate the proposed method with the real robot platform.
\subsection{Experiment Details}

We use an xArm 6 robotic arm with its two-finger gripper to conduct the experiment. In the system, the robot is directly linked with an RTX3060 6GB laptop, and we install a Realsense d435i at the end effector of the robot arm as the robot camera. The frequency of the robot camera is 7 Hz. The ZED 2 camera is connected to the robot computer and sends the vision data back to the user computer with a frequency of 30Hz.
For the normal grasping experiment, we only run the grasp pose control module on the robot computer. For the demonstration learning experiment, we run both grasps pose control and demonstration learning module in the same computer with a maximum batch size equal to 4. In the proposed demonstration learning module, the temperature parameter is 0.1.

\subsection{Evaluation Metric}
Similar to previous literature \cite{fang2023anygrasp,qin2020s4g}, we adopt two success rates to evaluate the performance. The attempt-centric success rate is defined as the fraction of the number of successful grasp attempts over the number of total grasp attempts. This metric measures the ability of the grasp solution in a single attempt.
The object-centric success rate is the ratio of the number of successfully grasped objects over the total number of objects and each object allows two attempts. The attempt-centric rate is more strict than the object-centric success rate.

\subsection{Normal Grasping}

In the normal grasping experiment, we set a workspace limit on a plane for safety. The detection pose is set to 0.45 meters above the plane. During the experiment, we randomly place the objects on the plane at a random orientation and take one picture. After that, the initial grasp pose generating module will generate initial grasp poses for all detected objects in the scene. The segmentation model only uses the RGB input to generate key points for objects, and the grasp pose control module can select key points and predict the grasp pose for the scene according to the segmentation results and the depth map. The initial grasp pose has the vertical direction and adjusted rotation on the yaw. The pre-grasp pose is 0.15 meters backward in the approaching direction.
Additionally, we collect over 200 objects from 78 categories. Similar to objects in any grasp \cite{fang2023anygrasp} and YCB dataset \cite{calli2015benchmarking}. The size of the objects ranges from $1.5 \times1.5 \times1.5 cm^3$ to $19.5 \times14.5 \times25 cm^3$. Fig \ref{objects} shows part of unknown grasping objects. These objects can be roughly divided into hardware, fruit, food, toy, and household. 
% Detailed information about the testing objects can be found in the Appendix.

\begin{table}
\centering
\small
\caption{Success rates of different methods on real normal grasp. "Dex." represents DexNet 4.0. "Any." represents Any Grasp. "Demo." means the result after learning demonstrations.
}
\setlength{\tabcolsep}{1mm}
\begin{tabular}{l|llll|llll}
\hline
\multirow{2}{*}{Object} & \multicolumn{4}{c|}{\begin{tabular}[c]{@{}c@{}}Attempt-Centric \\ Success Rate (\%)\end{tabular}} & \multicolumn{4}{c}{\begin{tabular}[c]{@{}c@{}}Object-Centric\\ Success Rate (\%)\end{tabular}} \\ \cline{2-9} 
                        & \multicolumn{1}{c|}{Dex}            & \multicolumn{1}{c|}{Ours} & \multicolumn{1}{c|}{Demo}           & \multicolumn{1}{c|}{Any}             & \multicolumn{1}{c|}{Dex}            & \multicolumn{1}{c|}{Ours}       & \multicolumn{1}{c|}{Demo}    & \multicolumn{1}{c}{Any}            \\ \hline
\multicolumn{1}{l|}{Hardware}                & \multicolumn{1}{l|}{59.4}           & \multicolumn{1}{l|}{78.0}  & \multicolumn{1}{l|}{89.3}          & 81.2            & \multicolumn{1}{l|}{97.2}           & \multicolumn{1}{l|}{97.7}    & \multicolumn{1}{l|}{100.0}       & \multicolumn{1}{l}{100.0}          \\ \hline
\multicolumn{1}{l|}{Ragdoll}                 & \multicolumn{1}{l|}{87.6}           & \multicolumn{1}{l|}{100.0}  & \multicolumn{1}{l|}{100.0}          & 100.0           & \multicolumn{1}{l|}{100.0}          & \multicolumn{1}{l|}{100.0}      & \multicolumn{1}{l|}{100.0}     & \multicolumn{1}{l}{100.0}          \\ \hline
\multicolumn{1}{l|}{Toy}                     & \multicolumn{1}{l|}{72.8}           & \multicolumn{1}{l|}{82.0}  & \multicolumn{1}{l|}{84.3}           & 92.9            & \multicolumn{1}{l|}{93.7}           & \multicolumn{1}{l|}{100.0}  & \multicolumn{1}{l|}{100.0}        & \multicolumn{1}{l}{100.0}          \\ \hline
\multicolumn{1}{l|}{Food\&Snack}                   & \multicolumn{1}{l|}{52.0}           & \multicolumn{1}{l|}{85.8}      & \multicolumn{1}{l|}{86.3}      & 98.3           & \multicolumn{1}{l|}{93.9}           & \multicolumn{1}{l|}{95.4}    & \multicolumn{1}{l|}{95.4}      & \multicolumn{1}{l}{100.0}          \\ \hline
\multicolumn{1}{l|}{Household}               & \multicolumn{1}{l|}{63.5}           & \multicolumn{1}{l|}{59.5}     & \multicolumn{1}{l|}{65.0}          & 84.5            & \multicolumn{1}{l|}{98.9}           & \multicolumn{1}{l|}{95.0}         & \multicolumn{1}{l|}{100.0}   & \multicolumn{1}{l}{100.0}          \\ \hline
% Fruits                  & \multicolumn{1}{l|}{-}              & \multicolumn{1}{l|}{96.1}            & -               & \multicolumn{1}{l|}{-}              & \multicolumn{1}{l|}{100.0}          & -              \\ \hline
\multicolumn{1}{l|}{All}                  & \multicolumn{1}{l|}{72.2}              & \multicolumn{1}{l|}{78.2}  & \multicolumn{1}{l|}{81.9}          & 93.3               & \multicolumn{1}{l|}{97.4}              & \multicolumn{1}{l|}{98.2}      & \multicolumn{1}{l|}{98.7}    & \multicolumn{1}{l}{100.0}             \\ \hline
\end{tabular}
\label{normalgrasp}
\end{table}

For each detection trial, the grasp pose control module segments the objects and provides an initial grasp pose in the scene within 80ms. After learning the human demonstration, the mean inference speed for each trial is less than 1s. The average robot trajectory planning time is 2.9s. The mean picks per hour (MPPH) mainly depends on the robot's moving speed and the gripper's executing speed. In the normal grasp experiment, the average pick speed is over 300 MPPH.
The experiment results in Table \ref{normalgrasp} show the superior grasping ability in some objects, especially for ragdolls and foods. Specifically, we achieved 100\% attempt-centric success rate on ragdolls. The attempt-centric successful rates on hardware (78\%), toy (82\%), and food (85.8\%) objects are also better than the supervised grasp solution Dexnet 4.0 \cite{mahler2019learning}. The performance on the household is not superior to the AnyGrasp \cite{fang2023anygrasp} because the segmentation model does not use the grasp pose dataset and the RGB-D images provided in the Graspnet \cite{fang2020graspnet}. Before demonstration information is applied, our self-supervised grasp pose-generating method lacks key information in grasping unseen objects.
% Detailed normal grasp results are shown in the Appendix.

\subsection{Grasping After Demonstration}

In this section, we choose the objects with low attempt-centric success rates in normal grasp experiments and demonstrate the grasping procedure through the AR-teleoperation system. To avoid the display error in the AR teleoperation system, we fine-tune the position of the last waypoints in the demonstration. The adjustment is based on the center point of the object. The number of waypoints in a single demonstration is less than 10 during the demonstration. The rotation threshold $t_r = 0.7$ and the translation threshold $t_l =0.9$. 
During the experiments, we first use the RGB-D input from the robot camera to segment objects in the scene and extract the point cloud feature. After that, the point cloud features are compared with the demonstrated point cloud features and generate the grasp pose. The contrastive point cloud learning model is pre-trained in the Shapenet \cite{chang2015shapenet} dataset. For each demonstrated object, we provide three demonstrations. 
Table \ref{demograsp} illustrates the attempt-centric successful rate for grasping results after learning the demonstration. The overall attempt-centric success rate increases from 78.3\% to 81.2\% after learning three demonstrations. In normal grasp experiment, it is difficult for the self-supervised model to grasp flat objects like plates and some small objects like screws and small screwdrivers. The framework learns the 6-DoF grasp pose from the demonstrated point clouds and waypoints. From the results, we also find that the grasping ability is enhanced when more demonstrations are provided.
The maximum robot planning time is 0.5 seconds.
The latency of teleoperation in robot manipulation is lower than 0.5 seconds.

\begin{table}
\centering
\small
\setlength{\tabcolsep}{1mm}
\caption{Attempt-Centric Successful Rate for 6-DoF grasp after learning different times of human demonstration. "No Demo" represents no human demonstration. }
\label{demograsp}
\begin{tabular}{c|clll}
\hline
\multirow{2}{*}{Objects} & \multicolumn{4}{c}{Attempt-Centric Success Rate(\%)}                                                                 \\ \cline{2-5} 
                        & \multicolumn{1}{c|}{No Demo} & \multicolumn{1}{c|}{1 Demo} & \multicolumn{1}{c|}{2 Demos} & \multicolumn{1}{c}{3 Demos} \\ \hline

\multicolumn{1}{l|}{Plates}  & \multicolumn{1}{c|}{0.0}     & \multicolumn{1}{c|}{70.0}       & \multicolumn{1}{c|}{90.0}        &  \multicolumn{1}{c}{90.0}                            \\ \hline
\multicolumn{1}{l|}{Bowls}  & \multicolumn{1}{c|}{20.0}     & \multicolumn{1}{c|}{70.0}       & \multicolumn{1}{c|}{90.0}        &  \multicolumn{1}{c}{90.0}                            \\ \hline
\multicolumn{1}{l|}{Mustard source}  & \multicolumn{1}{c|}{50.0}     & \multicolumn{1}{c|}{40.0}       & \multicolumn{1}{c|}{50.0}        &  \multicolumn{1}{c}{60.0}                            \\ \hline
\multicolumn{1}{l|}{Small screws}  & \multicolumn{1}{c|}{20.0}     & \multicolumn{1}{c|}{90.0}       & \multicolumn{1}{c|}{90.0}        &  \multicolumn{1}{c}{90.0}                            \\ \hline
\multicolumn{1}{l|}{Small screwdrivers}  & \multicolumn{1}{c|}{50.0}     & \multicolumn{1}{c|}{90.0}       & \multicolumn{1}{c|}{90.0}        &  \multicolumn{1}{c}{90.0}                            \\ \hline
\multicolumn{1}{l|}{All}  & \multicolumn{1}{c|}{78.2}     & \multicolumn{1}{c|}{80.6}       & \multicolumn{1}{c|}{81.2}        &  \multicolumn{1}{c}{81.9}                            \\ \hline
\end{tabular}
\end{table}

\begin{table}
\centering
\caption{Ablation study on contrastive point cloud learning model }
\small
\setlength{\tabcolsep}{1mm}
\begin{tabular}{l|cc|l}
\hline
     Method & RGB & Point Clouds & {\begin{tabular}[c]{@{}l@{}}Attempt-Centric \\ Success Rate (\%)\end{tabular}}  \\ \hline
No Demo & &  & \multicolumn{1}{c}{78.2}                             \\ 
\textbf{Case1} & & \checkmark & \multicolumn{1}{c}{\textbf{81.9}}                            \\ 
Case2 &\checkmark &  & \multicolumn{1}{c}{79.6}                             \\ 
Case3 &\checkmark & \checkmark & \multicolumn{1}{c}{80.6}                        \\ \hline
\end{tabular}
\label{ablation}
\end{table}

\subsection{Ablation Study}
In this section, we report the ablation study about the demonstration learning. Table \ref{ablation} illustrates the performance improvement after using point clouds contrastive learning (Case1), RGB image contrastive learning (Case2), and the RGB-D contrastive learning model (Case3). For comparison, we report the grasp performance without learning demonstrations.
The RGB images come from the ImageNet \cite{deng2009imagenet} dataset with the same category label. During training, we normalize the point clouds and use the same data augmentation methods.
We find that solely point clouds learning model (Case1) provides the most useful information and assists the system in generating 6-DoF grasp pose. Although RGB image contains morphological information, the point clouds may easily adapt the information to the adjustment of 6-DoF grasp pose.

\section{CONCLUSIONS}

This paper proposes a self-supervised 6-DoF robot grasping framework that efficiently learns human demonstration via an AR teleoperation system. In our framework, hidden morphology representation captured from point clouds can be efficiently learned by the contrastive learning model, which allows the robot to learn from human demonstrations. 
We apply the framework to the AR system to facilitate the visualization of robot grasping, demonstration recording, and demonstration learning in robot manipulation.
In future work, we are interested in extending the grasping ability of our system in a more dynamic environment.

%                                   % the textheight of the last page by a suitable amount.
%                                   % This command does not take effect until the next page
%                                   % so it should come on the page before the last. Make
%                                   % sure that you do not shorten the textheight too much.

%%%%%%%%%%%%%%%%%%%%%%%%%%%%%%%%%%%%%%%%%%%%%%%%%%%%%%%%%%%%%%%%%%%%%%%%%%%%%%%%

%%%%%%%%%%%%%%%%%%%%%%%%%%%%%%%%%%%%%%%%%%%%%%%%%%%%%%%%%%%%%%%%%%%%%%%%%%%%%%%%

\bibliographystyle{IEEEtran}

\bibliography{reference}

\end{document}